\documentclass[authoryear,review]{elsarticle}

\usepackage{amsmath,amsfonts,amssymb}
\usepackage{graphicx}
\usepackage{booktabs}
\usepackage[tight]{subfigure}
\usepackage{flushend}
\usepackage{rotating}
\usepackage{lineno}
\usepackage{units}
\begin{document}
\begin{frontmatter}
\title{Speckle Reduction with Adaptive Stack Filters}
\author{ Mar\'ia Elena Buemi\corref{cor1}}
\ead{mebuemi@dc.uba.ar}
\address{Departamento de Computaci\'on, Facultad de Ciencias Exactas y Naturales, Universidad de Buenos Aires,Pabell\'on I, Argentina}
\author{Alejandro C.\ Frery, Heitor S.\ Ramos}
\address{LCCV \& LaCCAN/CPMAT,	Universidade Federal de Alagoas, BR 104 Norte km 97,57072-970 Macei\'o, AL -- Brazil}
\cortext[cor1]{Corresponding author. Fax: +54 11 4576 3359}
\begin{abstract}
Stack filters are a special case of non-linear filters.
They have a good performance for filtering images with different types of noise while preserving edges and details.
A stack filter decomposes an input image into stacks of binary images according to a set of thresholds.
Each binary image is then filtered by a Boolean function, which characterizes the filter.
Adaptive stack filters can be computed by training using a prototype (ideal) image and its corrupted version, leading to optimized filters with respect to a loss function.
In this work we propose the use of training with selected samples for the estimation of the optimal Boolean function.
We study the performance of adaptive stack filters when they are applied to speckled imagery, in particular to Synthetic Aperture Radar (SAR) images.
This is done by evaluating the quality of the filtered images through the use of suitable image quality indexes and by measuring the classification accuracy of the resulting images.
We used SAR images as input, since they are affected by speckle noise that makes classification a difficult task.
\end{abstract}
\end{frontmatter}
\linenumbers
\section{Introduction}\label{introduc}

SAR images are generated by a coherent illumination system and are affected by the coherent interference of the signal from the terrain~\citep{OliverQuegan98}.
This interference causes fluctuations of the detected intensity which varies from pixel to pixel, an effect called speckle noise, that also appears in ultrasound-B, laser and sonar imagery.

Speckle noise, unlike noise in optical images, is neither Gaussian nor additive; it follows other distributions and is multiplicative.
Classical techniques, therefore, lead to suboptimal results when applied to this kind of imagery.
Among the authors that have studied the problem of adapting classical image processing methods to SAR data, \cite{Lee:1981:RFI} provides a good starting point.

The physics of image formation leads to the following model: the observed data can be described by the random field $Z$, defined as the product of two independent random fields: $X$, the backscatter, and $Y$, the speckle noise.

The backscatter is a physical magnitude that depends on the geometry and water content of the surface being imaged, as well as on the angle of incidence, frequency and polarization of the electromagnetic radiation emitted by the radar.
It is the main source of information sought in SAR data.

Speckle has a major impact on the accuracy of classification procedures~\citep{MejailJacoboFreryBustosIJRS,Capstick2001}, since it introduces a low signal-to-noise ratio.
The effectiveness of techniques for reducing speckle can be measured, among other quantities \citep[see][for instance]{Petty:LAAR:ProtocoloLee}, through the accuracy of simple classification methods.
The most widespread statistical classification technique is Gaussian maximum likelihood.

Different statistical distributions have been proposed in the literature for describing speckled data.
\citet{StatisticalModelingSARImagesSurvey} presents a comprehensive and updated survey of univariate distributions able to describe speckled data.
In this work, since we are dealing with intensity format, we use the Gamma distribution, denoted by $\Gamma$, for the speckle, and the reciprocal of Gamma distribution, denoted by $\Gamma^{-1}$, for the backscatter.
These assumptions, and the independence between the fields, result in the intensity $\mathcal{G}^{0}$  law for the return~\citep{frery96,MejailJacoboFreryBustosIJRS}.
This family of distributions is indexed by three parameters: roughness $\alpha$, scale $\gamma$, and the number of looks $L$, and it has been validated as an universal model for several types of targets.
Intensity data is obtained by summing the squared real and imaginary parts of the complex return; \citet{VasconcellosFrerySilva:CompStat} discuss properties of this type of speckled data.

Stack filters are a special case of non-linear filters.
They have good performance for improving images with different types of noise while preserving edges and details.
Various authors have studied these filters, and many methods have been developed for their construction and applicaton~\citep{Prasad2005,Shi2005}.

These filters decompose the input image, by thresholds, in binary slices forming a stack of data.
Each binary image is then filtered using a Boolean function evaluated on a sliding window.
The resulting image is obtained summing up all the filtered binary images.
The application of stack filters to speckled data was studied by~\citet{Buemi:Sibgrapi:2007,SARFilterStackCIARP2011}.

The main drawback of using stack filters is the need to compute the Boolean functions that satisfy a certain criterion.
Direct computation on the set of all Boolean functions is unfeasible, and promising techniques rely on a learning procedure: the use of a pair of images, namely the ideal and corrupted one, and of a loss function.
The Boolean functions are sought to provide the best estimator of the former using the latter as input.
The stack filter design method used in this work is based on an algorithm proposed by \citet{YooKelvinHuangCoyleAdams}.
The drawback of this line of action is the need of a pure, noiseless image.

\citet{EchocardiographicSpeckleReductionComparison} provide a comprehensive and updated review of the literature on speckle filters, with a view towards echocardiographic imagery.
They propose a categorization, within which our proposal should be included in the ``SAR'' category.
 
We propose the use of user-provided information.
The user selects as many regions of interest as desired, and after a descriptive and quantitative analysis of the data, he/she provides an ``ideal'' value for each region.
The Boolean functions are then sought to provide an estimator of such values in the corresponding areas.
This approach reduces the computational effort of building the Boolean functions and, at the same time, gives the option of providing a noiseless complete image or specifying ``ideal'' values in regions chosen by the user.

%%% ACF Cuidado acá, hay que decir lo que, de hecho, se usó
%%% MEB En este trabajo,"construimos" filtros stack adaptativos a partir de imagenes SAR simuladas y reales. Evaluamos los indices de calidad de imagen, sobre imágenes simuladas comparando con el filtro de Lee. Sobre imágenes SAR reales clasificamos por MV y utilizamos los filtros de Lee y Frost para cuantificar la tasa de clasificación, además incorporamos el índice de preservación de contraste sobre imágenes simuladas con gran cantidad de bordes y superficies de distintos tamaños, este índice fue obtenido para los filtros Stack, Lee y Frost. 
%%% MEB aqui es donde tenemos inconsistencia porque aparece Kuan pero no lo pusimos, entonces, calcule los indices de contraste para Kuan de las tres filas de la tabla 3 y agregue una imagen pdf en la carpeta pdf
We study the application of this type of filter to SAR images, assessing its performance by evaluating the quality of the filtered images through the use of objective image quality indexes like the universal image quality index and the correlation measure index and by measuring the classification accuracy of the resulting images using maximum likelihood Gaussian classification.

The structure of this paper is as follows.
Section~\ref{sec:imagSAR} summarizes the $\mathcal{G}^{0}$ model for speckled data. 
Section~\ref{define_stack} reviews stack filters, and describes the filter design method used in this work. 
In Section~\ref{sec:Resultados} we discuss the results of filtering through image quality assessment and classification performance.
Finally, in Section~\ref{sec:Conclusiones} we present the conclusions.

\section{The Multiplicative Model}\label{sec:imagSAR}

Following~\citet{Petty:LAAR:ProtocoloLee}, we will only present the univariate intensity case.
Other formats (amplitude and complex) are treated in detail in~\cite{frery96}.

The intensity $\mathcal G^0$ distribution that describes speckled return $Z$ is characterized by the following density:
\begin{equation*}
f(z)=\frac{n^{n}\Gamma(n-\alpha)}{\gamma^{\alpha}\Gamma(n)\Gamma(-\alpha)}\frac{z^{n-1}}{(\gamma+nz)^{n-\alpha}},
\end{equation*}
where $-\alpha,\gamma,z>0$, $n\geq1$.
This situation is denoted $Z\sim\mathcal G^{0}(\alpha,\gamma,n)$.

The $\alpha$ parameter describes the image roughness or texture.
It adopts negative values, varying from $-\infty$  to $0$.
If $\alpha$ is near $0$, then the image data are extremely rough (for example: urban areas), and if $\alpha$ is far from the origin then the data correspond to a smooth region (for example: pasture areas). 
The values for forests lay in-between.

The $r$-th order moment of a $\mathcal G^{0}(\alpha,\gamma,n)$-distributed random variable is given by
\begin{equation}
\operatorname{E}(Z^r) = \Bigl( \frac{\gamma}{n}\Bigr)^r \frac{\Gamma(-\alpha-r)\Gamma(n+r)}{\Gamma(-\alpha)\Gamma(n)},
\label{eq:GI0Moments}
\end{equation}
if $-n>\alpha$ and infinite if otherwise.

Many filters have been proposed in the literature for reducing speckle noise, among them the ones by~\citet{Lee86}, by~\citet{kuan87} and by~\citet{FrostFilter}.
These filters will be applied to speckled data, along with the filter proposed in this work. 
%%% ACF ===>>> Acá hay que ordenar <<<===
For quality performance the comparison will be done between the results of applying the stack filter and the Lee filter, since the latter is considered one of the touchstones for speckle reduction.
Classification performance will be assessed by classifying data filtered with the Lee, Frost and stack filters using Gaussian maximum likelihood.
The two first filters can be considered classical choices in the area.

\section{Stack Filters}\label{define_stack}

This section is dedicated to a brief synthesis of stack filter definitions and design.
For more details on this subject, the reader is referred to the works by \citet{AstolaKuosmanen,LinKim,YooKelvinHuangCoyleAdams}.

Consider images of the form $X\colon S\rightarrow \{0,\dots,M\}$, with $S$ the support and $\{0,\dots,M\}$ the set of admissible values.
The threshold is the set of operators $T^{m}\colon \{0,\dots,M\}\rightarrow\{0,1\}$ given by 
\begin{equation*}
T^{m}(x)=\left\{ 
\begin{array}{ccc}
1 & \text{if} & x\geq m, \\ 
0 & \text{if} & x < m.
\end{array}
\right.   \label{def_umbral}
\end{equation*}
We will use the notation
$X^{m}=T^{m}(x)$.  %\label{form_reconstruccion}
According to this definition, the value of a non-negative integer number $x\in \{0,\dots,M\}$ can be reconstructed making the summation of its thresholded values between $0$ and $M$. 

In the following, we show an example of the threshold decomposition of an unidimensional signal.
Let $X=[2,1,3,2,3]$ be a signal.
Its decomposition is given by:
$X^1 = [1,1,1,1,1]$,
$X^2 = [1,0,1,1,1]$,
$X^3 = [0,0,1,0,1]$.

Let $X=(x_{0},\ldots,x_{n-1})$ and $Y=(y_{0},\ldots,y_{n-1})$ be binary vectors of length $n$.
Define an order relation given by
$X\leq Y$ if and only if $x_{i}\leq y_{i}$ holds true for every $i$.
This relation is reflexive, anti-symmetric and transitive, generating therefore a partial ordering on the set of binary vectors of fixed length.

A Boolean function $f\colon\{ 0,1\}^{n}\rightarrow\{0,1\}$, where $n$ is the length of the input vectors, has the stacking property if and only if 
\begin{equation*}
\forall X,Y\in\{0,1\}^{n},\ X\leq Y\Rightarrow f(X)
\leq f( Y) .  \label{prop_stacking}
\end{equation*}

We say that $f$ is a positive Boolean function if and only if it can be written by means of an expression that contains only non-complemented input variables.
That is,
\begin{equation}
f( x_{1},x_{2},\ldots ,x_{n})=\bigvee_{i=1}^{K} \bigwedge_{j\in P_{i}}x_{j},
\label{form_func_bool_positiva}
\end{equation}%
where $n$ is the number of arguments of the function, $K$ is the number of terms of the expression and $P_{i}$ is a subset of the interval $\{1, \ldots, N\}$, `$\vee$' and `$\wedge$' denote, respectively, the AND and OR Boolean operators.
It is possible to proof that this type of functions has the stacking property.

A stack filter is defined by the function $S_{f}\colon \{0, \ldots, M\}^{n}\rightarrow \{0, \ldots, M \} $, corresponding to the Positive Boolean function $f(x_{1}, x_{2}, \ldots, x_{n}) $ expressed in the form given in equation~(\ref{form_func_bool_positiva}).
The function $S_{f}$ can be expressed by means of a summation:
\begin{equation*}
S_{f}( X) =\sum_{m=1}^{M}f( T^{m}( X)) .
\label{form_reconstruccion_equivalencia}
\end{equation*}

In this work we applied the stack filter generated with the fast algorithm described in \citet{LinSellkeCoyle,LinKim,YooKelvinHuangCoyleAdams}.

Stack filters are built by a training process that generates a positive Boolean function that preserves the stacking property.
Originally, this training is performed providing two complete images on $S$, one degraded and one noiseless.
The algorithm seeks the operator that best estimates the later using the former as input, with respect to a loss function.

%%% ACF Qué quiere decir la primera oración?
%%% MEB: La función booleana hallada en la etapa de entrenamiento, el filtro stack, contiene información acerca de la verdad en cada punto, por lo que es posible "reaplicarla" reiteradamente a la imagen ya filtrada.
The implementation developed for this work supports the application of the stack filter many times. 
Our approach consists of using, instead of that pair of images, a set of regions of interest, much smaller than the whole data set, and relying on the analysis the user makes of this information.
Besides not needing a noiseless image, the user is free to impose its prior knowledge and assumptions on the resulting image.

The performance of the proposed filters is assessed both by qualitative and quantitative analyses. 
The system has an interface which shows the user the mean value of each region, and suggests it as the default ``desired'' value, but he/she can choose other from a menu (including the median, the lower and upper quartiles and a free specification).
This freedom of choice is particularly useful when dealing with non-Gaussian degradation as is the case of, for instance, impulsive noise.

\section{Results}\label{sec:Resultados}

In this section, we present the results of building stack filters by the aforementioned training. 
These filters are applied to both simulated and real data. 
The quality of the results is assessed in two different manners: one using image quality objective measures, and other evaluating the influence of filtering on maximum likelihood classification.

\subsection{Image quality indexes}\label{sec:CalidadDeImagen}

The indexes used to evaluate the quality of the filtered images are the universal image quality index \cite{WangBovik}  and the correlation measure $\beta$.
The universal image quality index $Q$ is given by equation (\ref{form_WangBovikIndex})
\begin{equation}
Q = \frac{\sigma_{XY}}{\sigma_{X} \sigma_{Y}} \frac{2\overline{X}\,\overline{Y}}{\overline{X}^2+\overline{Y}^2}
\frac{2\sigma_{X}\sigma_{Y}}{\sigma_{X}^2+\sigma_{Y}^2},
\label{form_WangBovikIndex}
\end{equation}
where $\sigma_{X}^2 = (N-1)^{-1}\Sigma_{i=1}^N (X_i - \overline{X})^2$, $\sigma_{Y}^2 = (N-1)^{-1}\Sigma_{i=1}^N (Y_i - \overline{Y})^2$, $\overline{X} = N^{-1}\Sigma_{i=1}^N X_i$ and $\overline{Y} = N^{-1}\Sigma_{i=1}^N Y_i$.
The dynamic range of index $Q$ is $[-1,1]$, being $1$ the best value. To evaluate the index of the whole image, local indexes $Q_i$ are calculated for each pixel using a suitable square window, and then these results are averaged to yield the total image quality $Q$.
The correlation measure is given by equation (\ref{form_Beta})
\begin{equation}
\beta = \frac{\sigma_{\nabla^{2}X\nabla^{2}Y}}{\sigma^2_{\nabla^{2}X} \sigma^2_{\nabla^{2}Y}}, 
\label{form_Beta}
\end{equation}
where $\nabla^{2}X$ and $\nabla^{2}Y$ are the Laplacians of images $X$ and $Y$, respectively. 
The correlation measure range is $[-1,1]$.

\subsection{Observed metrics}\label{sec:ObservedMetrics}

A Monte Carlo experiment was performed, generating 1000 independent replications of synthetic 1-look SAR images for each of four contrast ratios. 
The generated images consist of two regions separated by a vertical straight border. 
Each sample corresponds to a different contrast ratio, which ranges from $10$:$1$ to $10$:$8$. 
This was done in order to study the effect of the contrast ratio in the quality indexes considered.

Table~\ref{tab:imagequality} shows the mean correlation measure $\beta$ and the mean quality index $Q$, as computed in the Monte Carlo experiment.  
The comparison is made between Lee filtered and stack filtered SAR images. 

\begin{table}[hbt]
\caption{Statistics from image quality indexes}\label{tab:imagequality}
    \centering
    \begin{tabular}{ crrrr}\toprule
	   & \multicolumn{4}{c}{$\beta$ index} \\ \cmidrule(lr{5pt}){2-5}
	       &       \multicolumn{2}{c}{Stack filter}  & \multicolumn{2}{c}{Lee filter} \\
	       \cmidrule(lr{5pt}){2-3} \cmidrule(lr{5pt}){4-5}
	    contrast & \multicolumn{1}{c}{$\overline{\beta}$} & \multicolumn{1}{c}{$s_\beta$} &     \multicolumn{1}{c}{$\overline{\beta}$} & \multicolumn{1}{c}{$s_\beta$}   \\ \midrule
	    $10:1$	& $0.1245$ & $0.0156$  & $0.0833$ & $0.0086$\\
	    $10:2$	& $0.0964$ & $0.0151$  & $0.0663$ & $0.0079$\\
	    $10:4$	& $0.0267$ & $0.0119$  & $0.0421$ & $0.0064$\\
	    $10:8$	& $-0.0008$ & $0.0099$ & $0.0124$ & $0.0064$\\ \midrule
	  &  \multicolumn{4}{c}{$Q$ index} \\ \cmidrule(lr{5pt}){2-5}
	       &       \multicolumn{2}{c}{Stack filter}  & \multicolumn{2}{c}{Lee filter} \\ 
	       	       \cmidrule(lr{5pt}){2-3} \cmidrule(lr{5pt}){4-5}
	    contrast & \multicolumn{1}{c}{$\overline{Q}$} & \multicolumn{1}{c}{$s_Q$} &     \multicolumn{1}{c}{$\overline{Q}$} & \multicolumn{1}{c}{$s_Q$}   \\ \midrule
	    $10:1$	& $0.0159$ & $0.0005$ & $0.0156$ & $0.0004$     \\
	    $10:2$	& $0.0154$ & $0.0005$ & $0.0148$ & $0.0004$     \\
	    $10:4$	& $0.0124$ & $0.0008$ & $0.0120$ & $0.0006$     \\
	    $10:8$	& $0.0041$ & $0.0013$ & $0.0021$ & $0.0006$     \\ \bottomrule
        \end{tabular}
\end{table}

It can be seen that, according to the results obtained for the $\beta$ index, the stack filter exhibits a better performance at high contrast ratios, namely $10$:$1$ and $10$:$2$, while the Lee filter shows the opposite behavior.
The results for the $Q$ index show slightly better results for the stack filter all over the range of contrast ratios. It is remarkable the small variance of these estimations, compared to the mean values obtained.

\begin{sidewaysfigure}[htb]
\centering
\subfigure[Values of $\beta$, Lee filter\label{fig:betaLee}]{\includegraphics[width=0.4\linewidth]{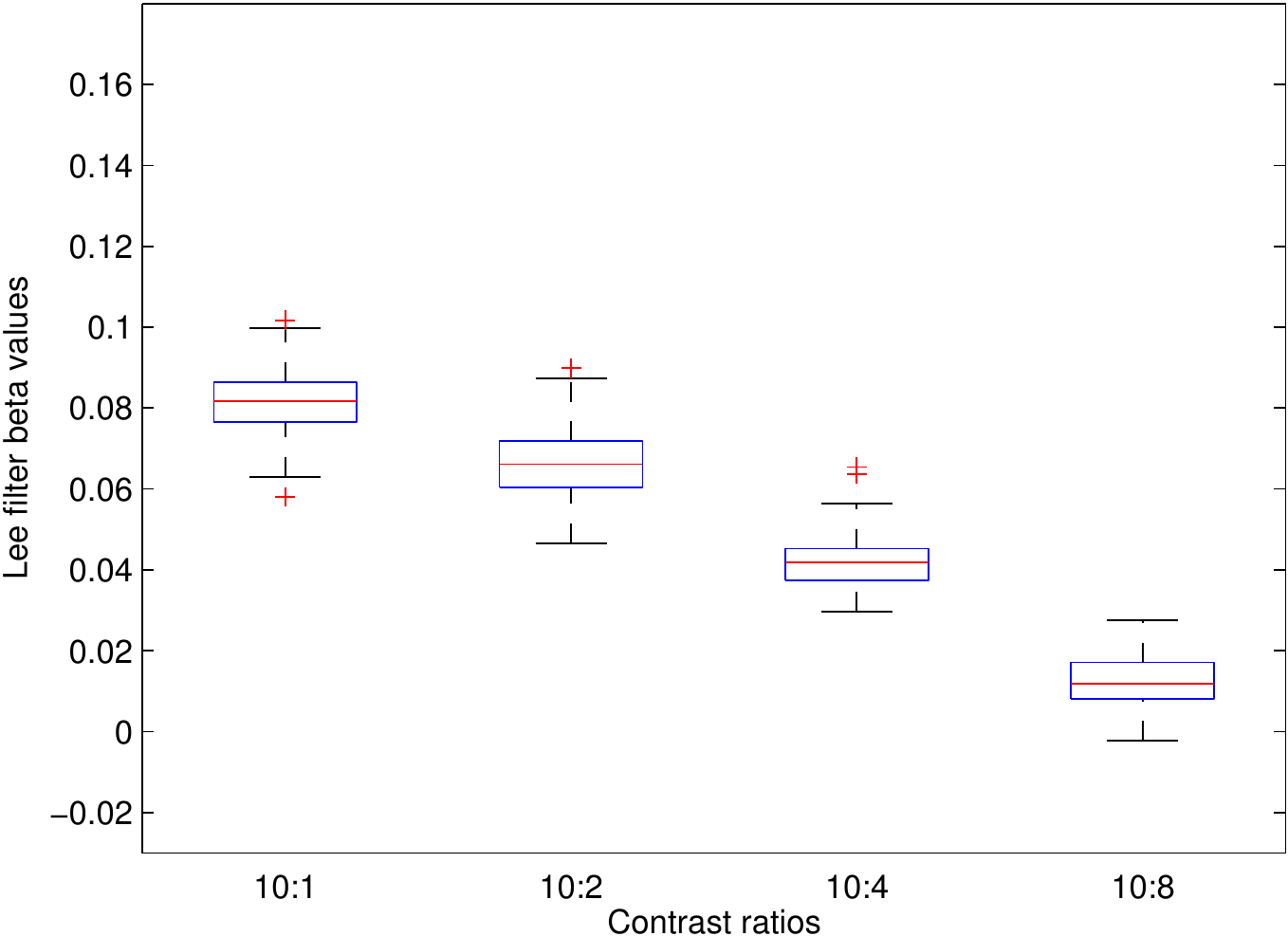}} 
\subfigure[Values of $\beta$, Stack filter\label{fig:betaStack}]{\includegraphics[width=0.4\linewidth]{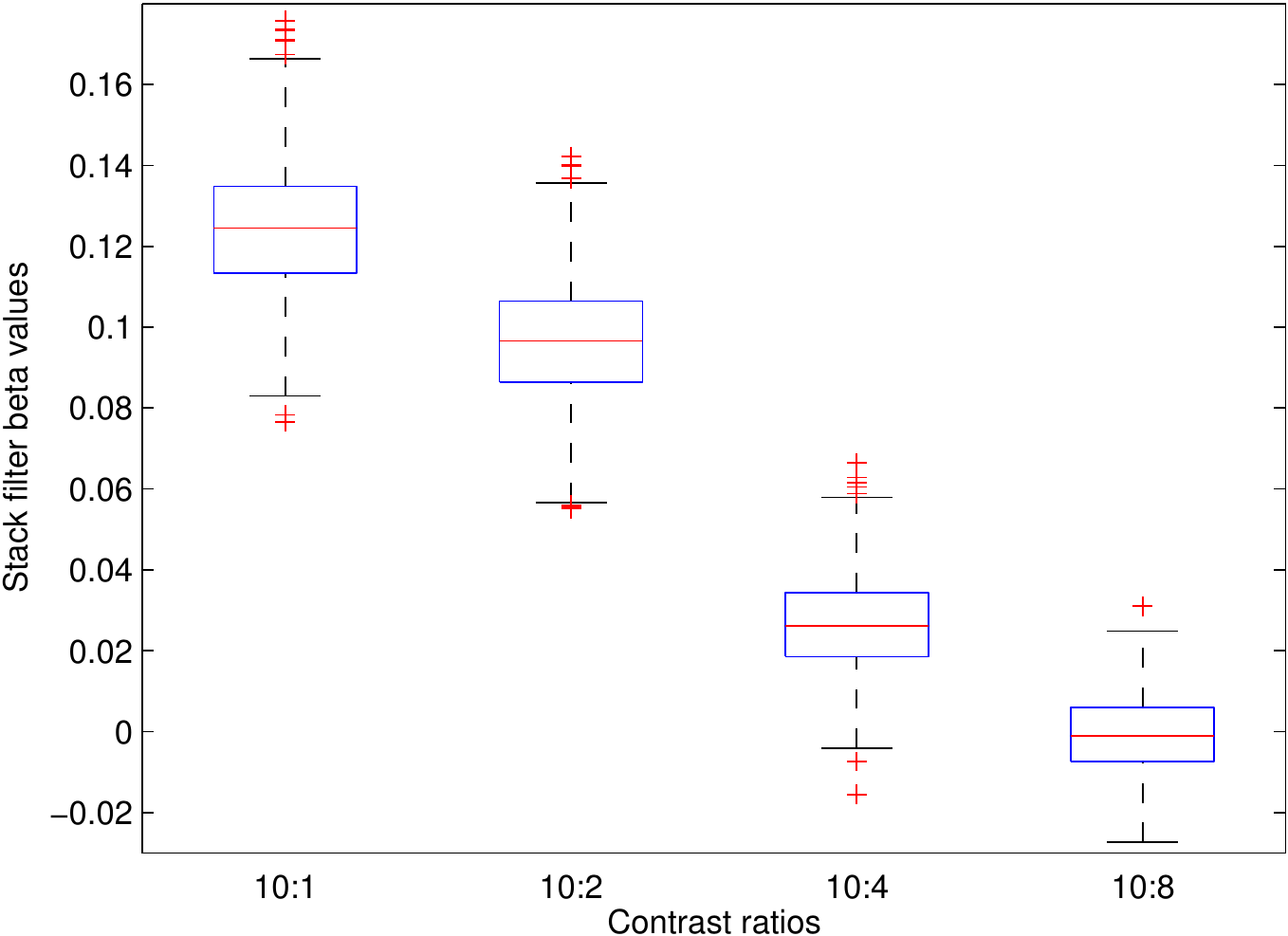}}
\subfigure[Lee filter values of $Q$\label{fig:QLee}]{\includegraphics[width=0.4\linewidth]{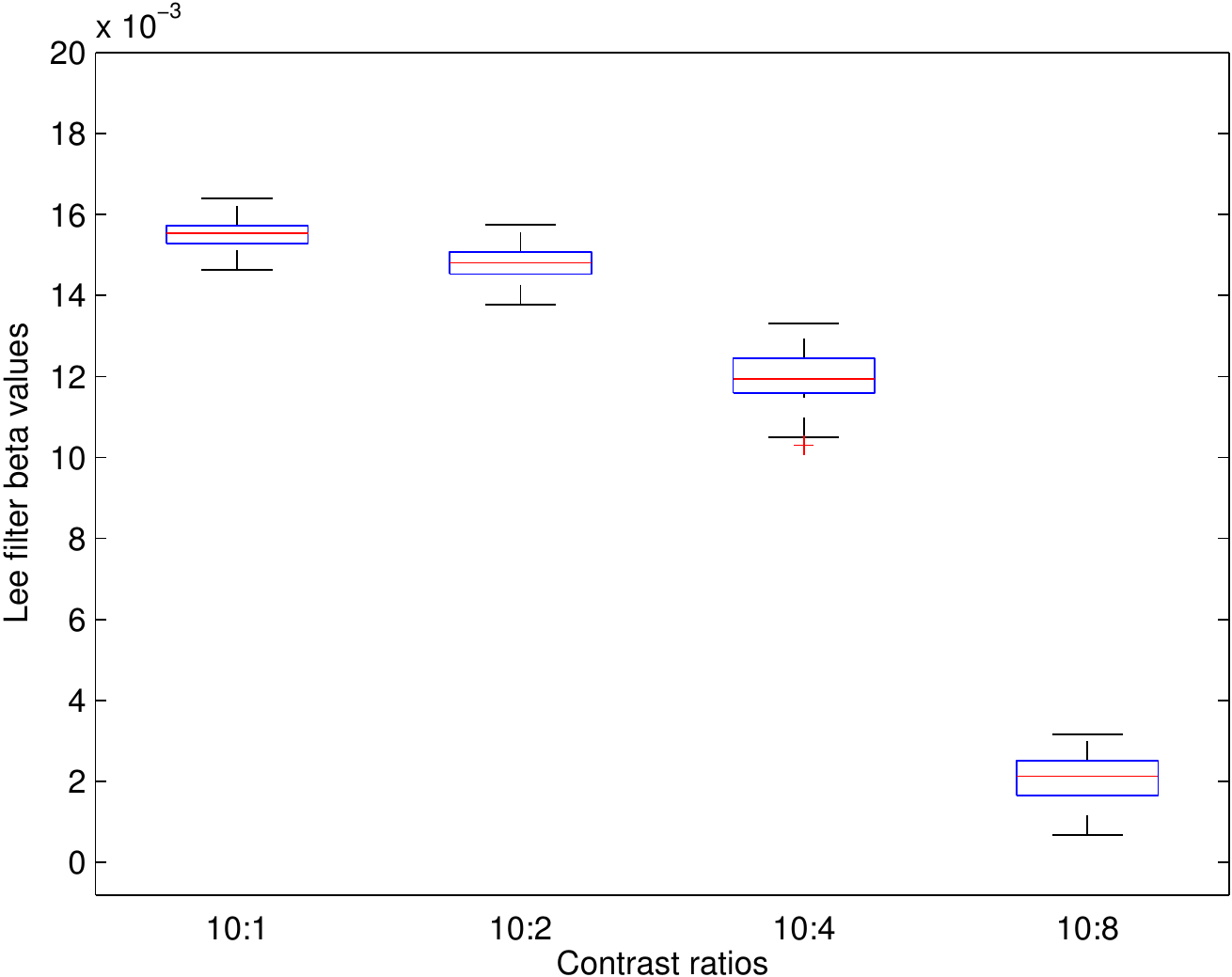}}
\subfigure[Stack filter values of $Q$\label{fig:QStack}]{\includegraphics[width=0.4\linewidth]{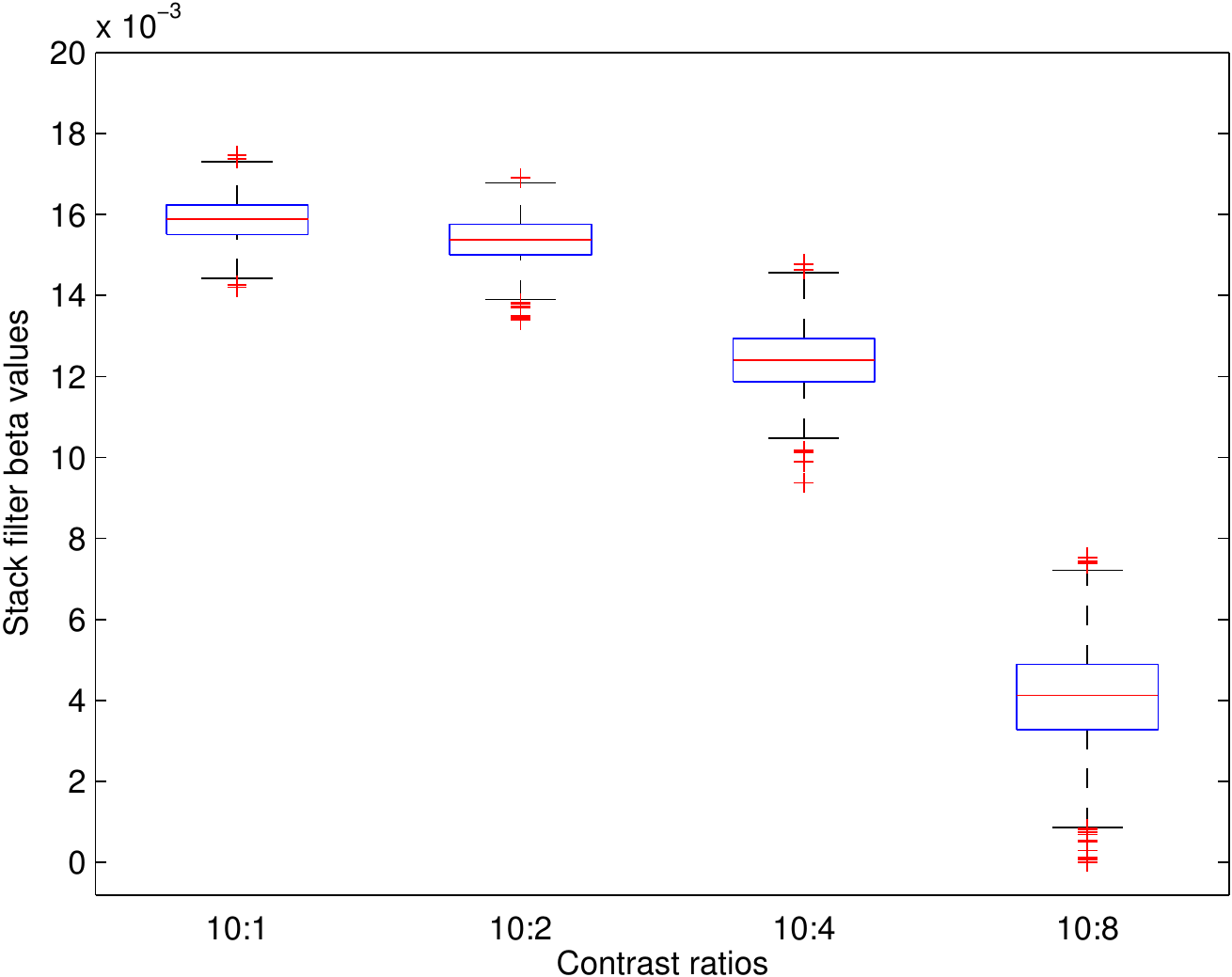}}
\caption{Boxplots of the quality indexes}
\label{fig:boxplots}
\end{sidewaysfigure}

Figure~\ref{fig:boxplots} shows the boxplots of the observations summarized in Table~\ref{tab:imagequality}. From the plots for the $\beta$ index, it can be seen that, the Lee filter has a lower degree of variability with contrast and that both are almost symmetric.  The plots of the $Q$ index show a better performance for the stack filter for all the contrast ratios considered.

The image resulting of applying the stack filter 95 times and the image produced by the application of the Lee filter. 
This was found the ideal number of iterations for this study: less iterations produced images with visible noise, while more iterations introduced blurring.
An automatic stopping criterion is a venue for future research.
The presented results are the mean values obtained from a Monte Carlo experiment involving different contrast ratios.

\subsection{Classification performance}\label{sec:SegmentationPerformance}

The equality of the classification results are obtained by calculating the confusion matrix, after Gaussian Maximum Likelihood Classification (GMLC).

\begin{figure*}[hbt]
\centering
\subfigure[Simulated image and GMLC\label{fig:simulated2reg}]{\includegraphics[width=.450 \linewidth]{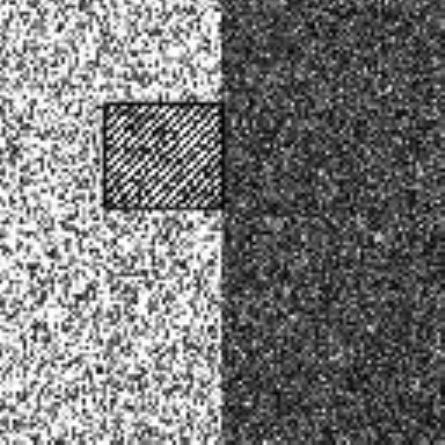}
\includegraphics[width=.45 \linewidth]{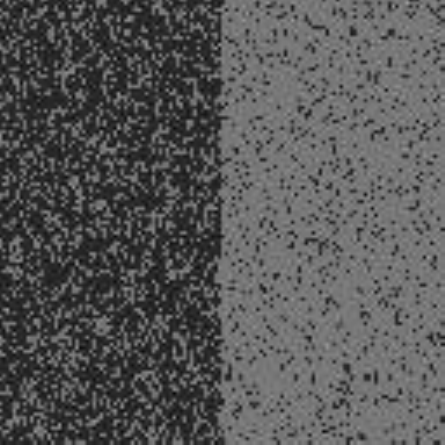}} \\
\subfigure[One iteration and GMLC\label{fig:1itersimul}]{ \includegraphics[width=.450 \linewidth]{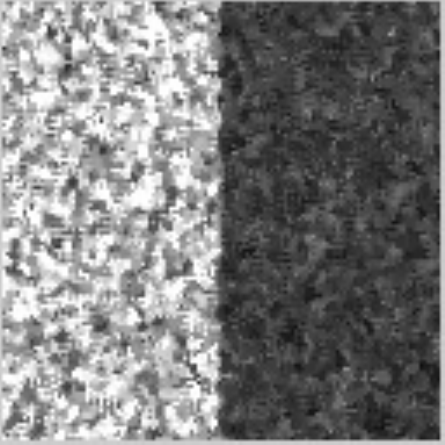}
\includegraphics[width=.45 \linewidth]{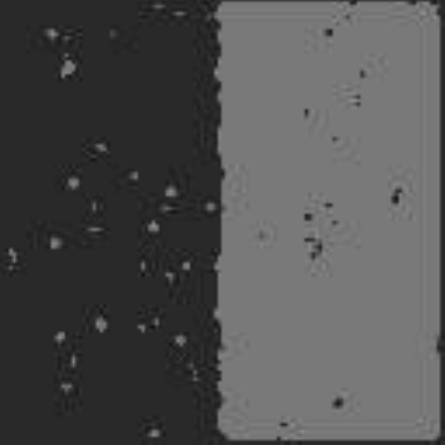}}\\
\subfigure[$95$ iterations and GMLC\label{fig:95itersimul}]{\includegraphics[width=.450 \linewidth]{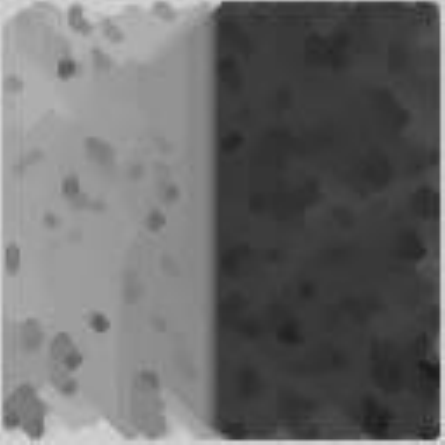}
\includegraphics[width=.45 \linewidth]{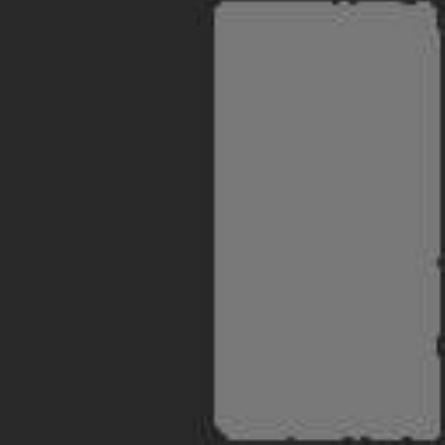}} 
 \caption{Training by region of interest: simulated data}
\label{fig:training_simulated}
\end{figure*}

Figure~\ref{fig:simulated2reg}, left, presents an image $128\times128$ pixels, simulated with two regions: samples from the $\mathcal G^0(-1.5,\gamma^*_{-1.5,1},1)$ and from the $\mathcal G^0(-10,10\gamma^*_{-10,1},1)$ laws form the left and right halves, respectively, where $\gamma^*_{\alpha,n}$ denotes the scale parameter that, for a given roughness $\alpha$ and number of looks $n$ yields an unitary mean law.
In this manner, Figure~\ref{fig:simulated2reg} presents data that are hard to classify: extremely heterogeneous and homogeneous areas with the lowest possible signal-to-noise ratio ($n=1$).
The mean value of the dashed area was used as the ``ideal'' image.
Figures~\ref{fig:1itersimul} and~\ref{fig:95itersimul}, left, show the result of applying the resulting filter once and $95$ times, respectively.
The right side of Figures~\ref{fig:simulated2reg}, \ref{fig:1itersimul} and~\ref{fig:95itersimul} present the GMLC of each image.
Not only the pointwise improvement is notorious, but the edge presevation is also noteworthy, specially in Figure~\ref{fig:95itersimul}, right, where the straight border has been completely retrieved.

\begin{sidewaysfigure}[htb]
\centering
 \subfigure[Image, samples and GMLC\label{fig:realimage}]{ \includegraphics[width=.24 \linewidth]{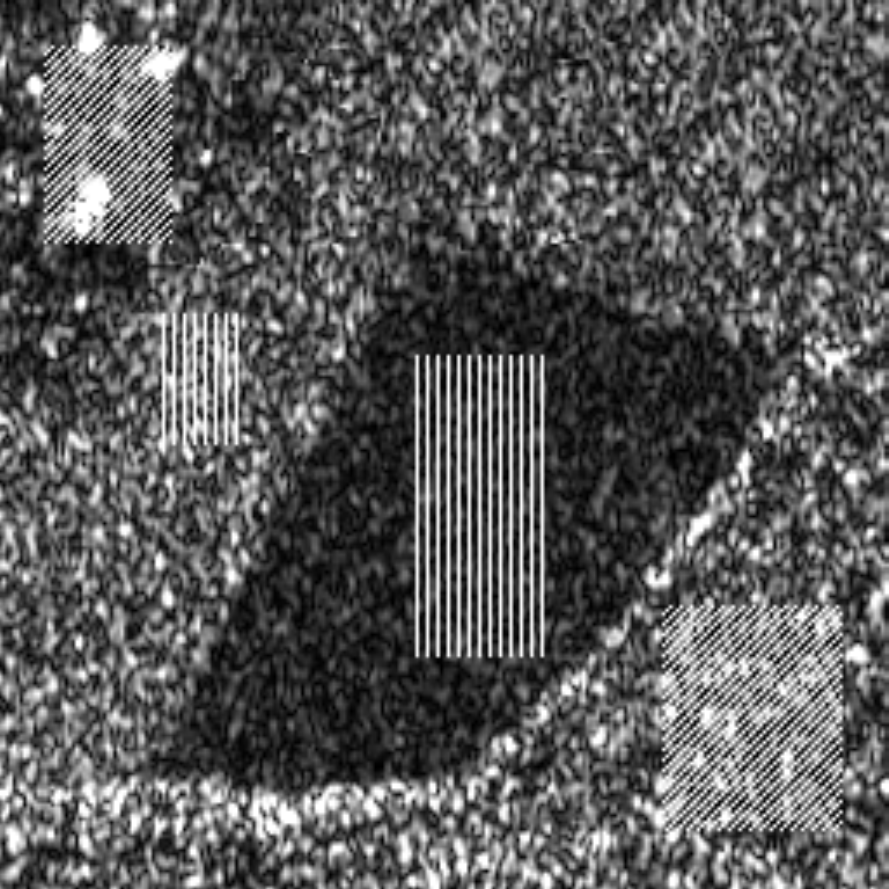}
\includegraphics[width=.24 \linewidth]{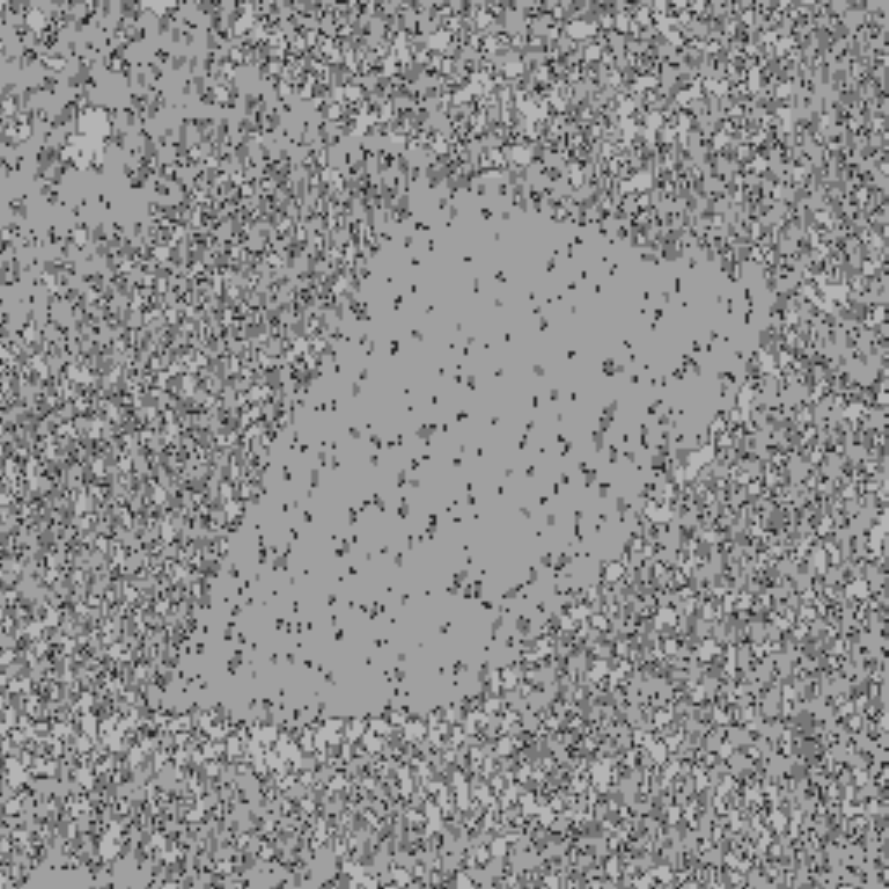}}
 \subfigure[Frost and GMLC\label{fig:filfrost}]{\includegraphics[width=.24 \linewidth]{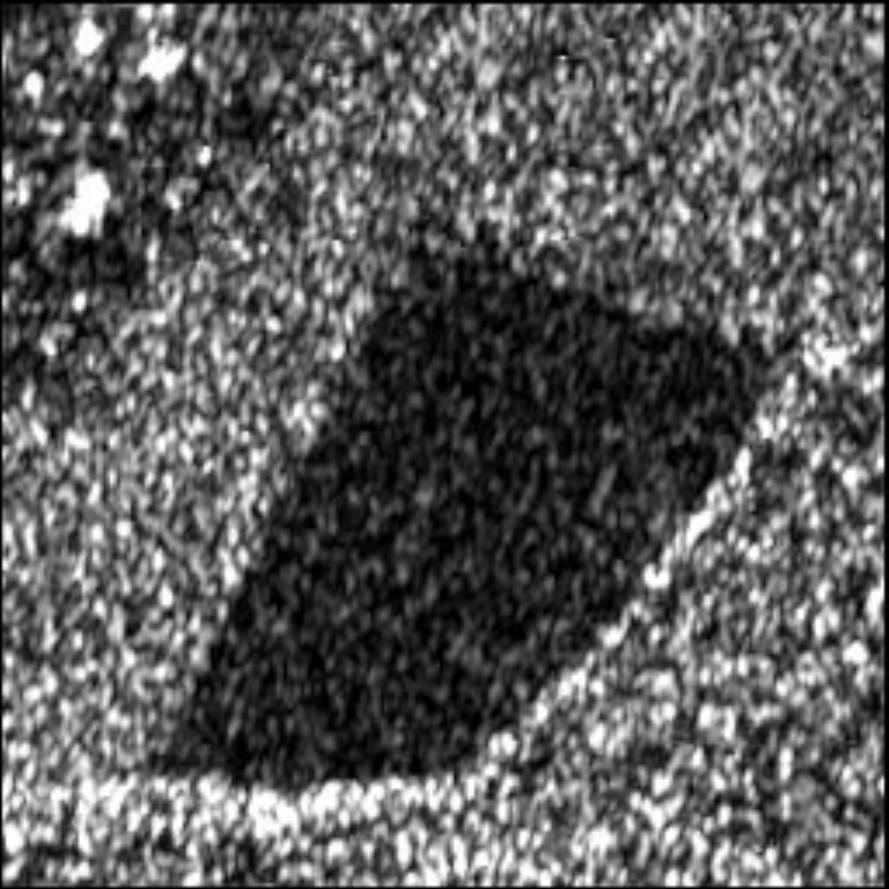}
\includegraphics[width=.24\linewidth]{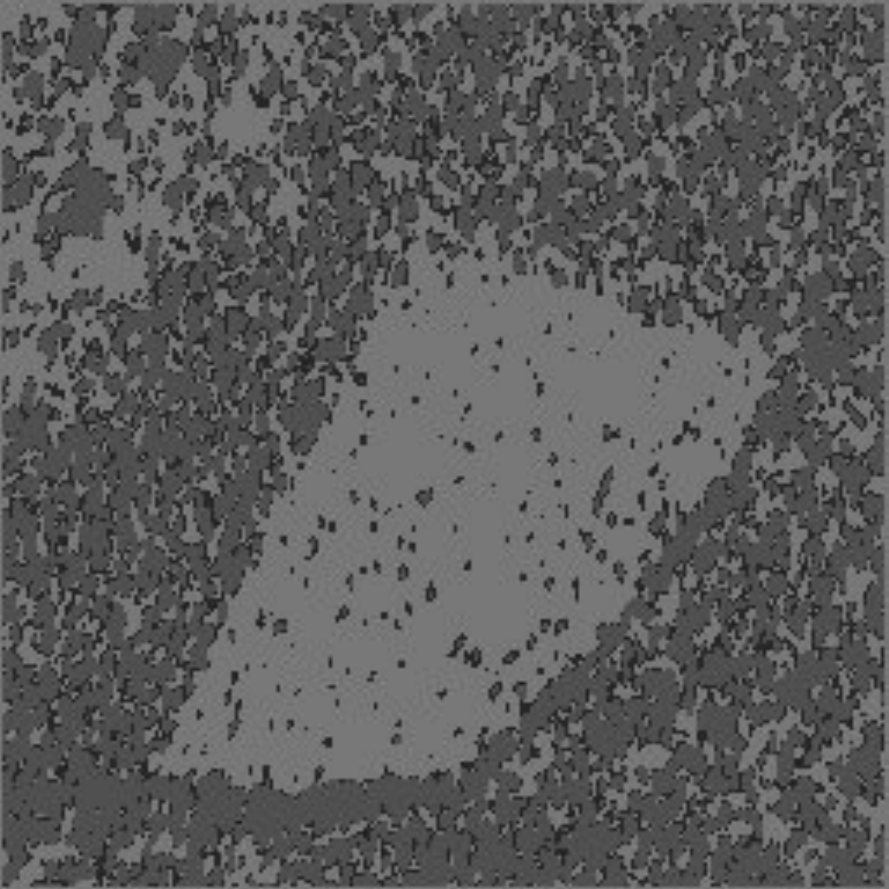}}\\
 \subfigure[Lee and GMLC\label{fig:fillee}]{\includegraphics[width=.24 \linewidth]{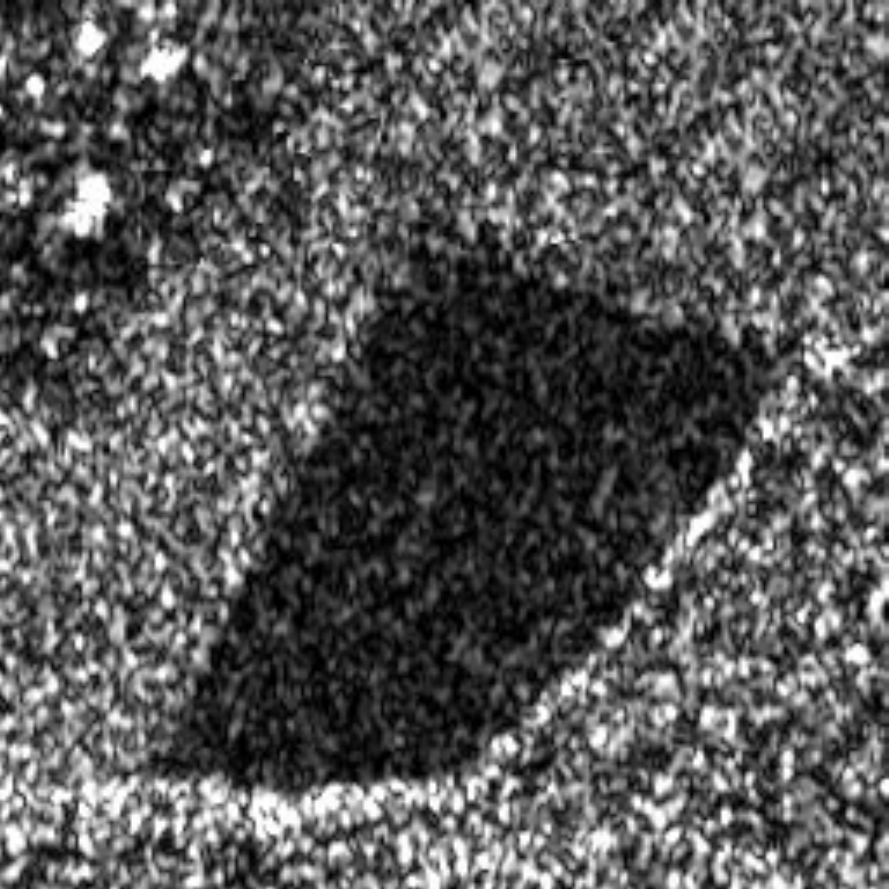}
 \includegraphics[width=.24 \linewidth]{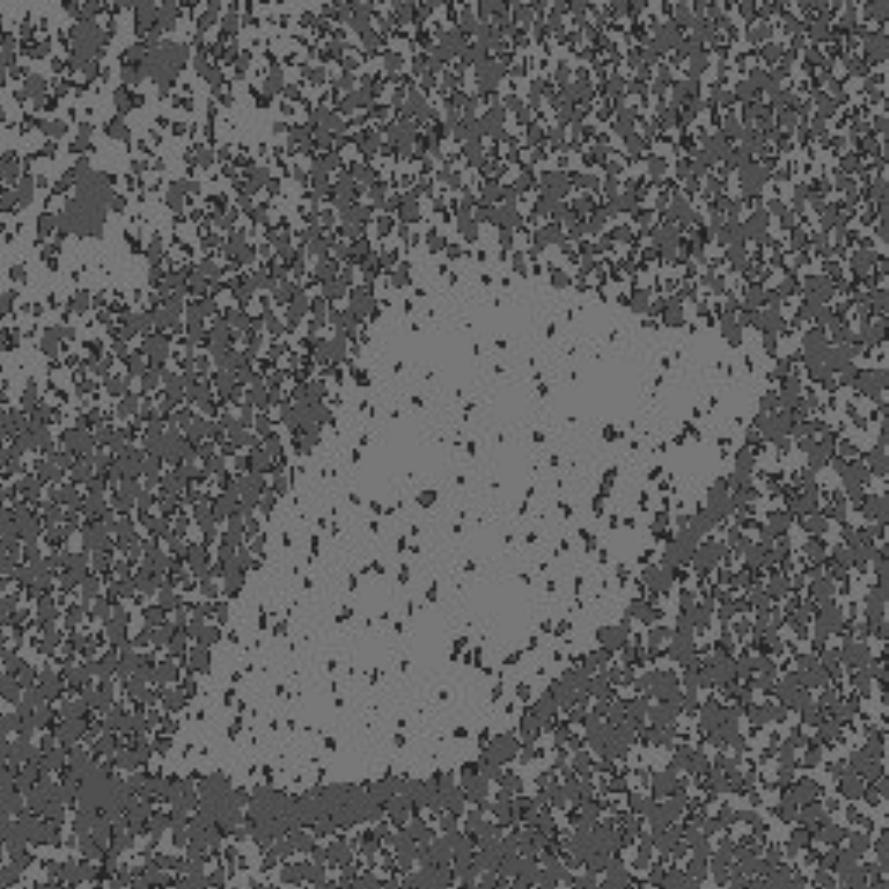}}
 \subfigure[Stack Filter $1$ and GMLC\label{fig:stackfilter1}]{\includegraphics[width=.24 \linewidth]{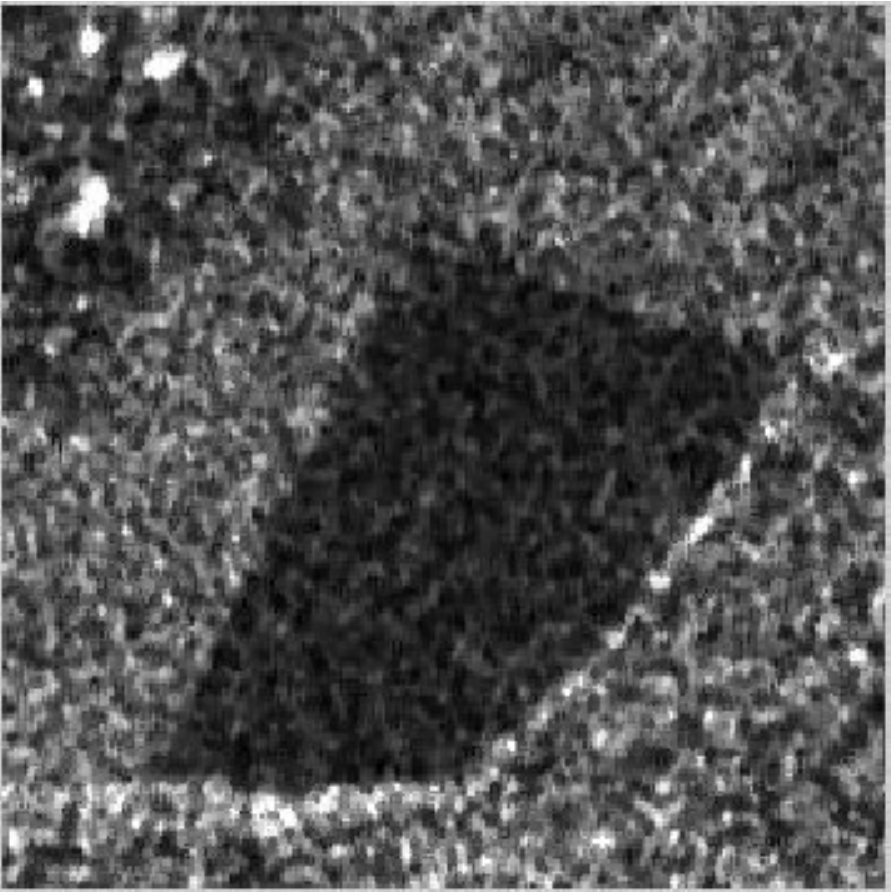}
 \includegraphics[width=.24 \linewidth]{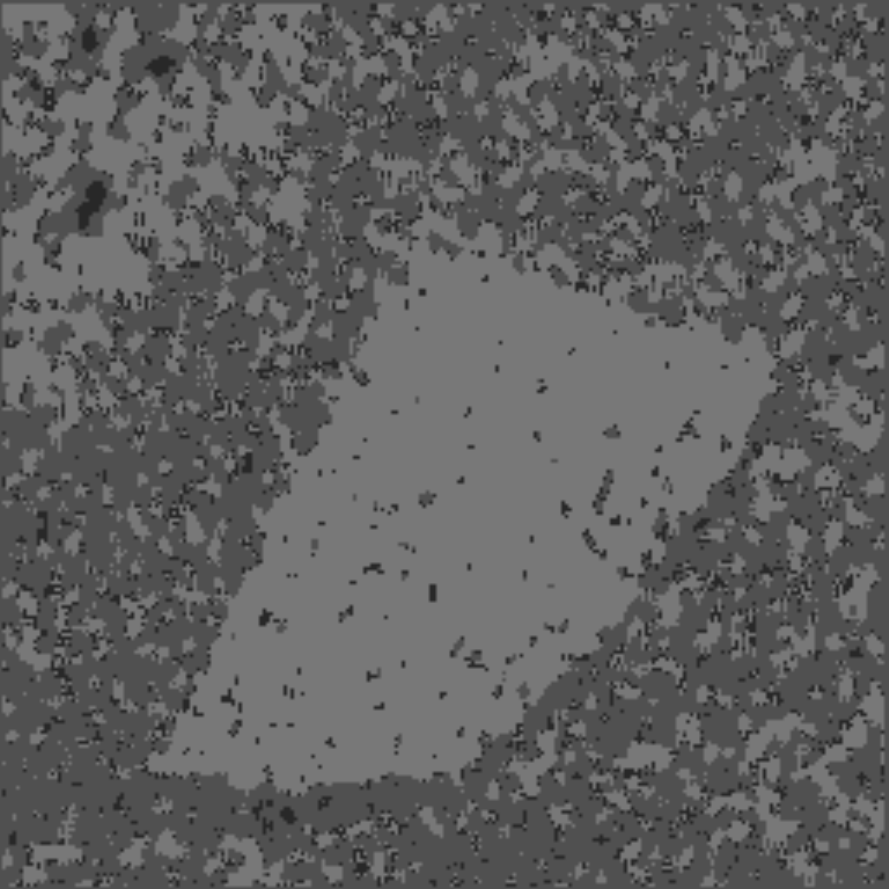}} \\
\subfigure[Stack Filter $20$ and GMLC \label{fig:stackfilter95}]{\includegraphics[width=.24\linewidth]{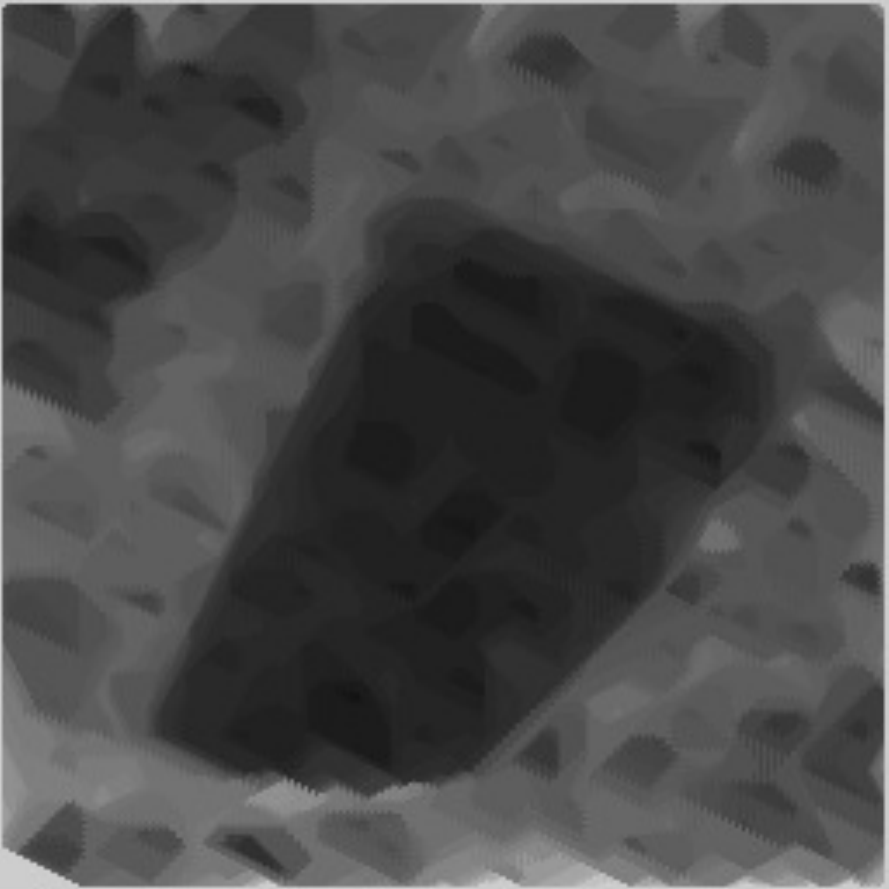}
\includegraphics[width=.24\linewidth]{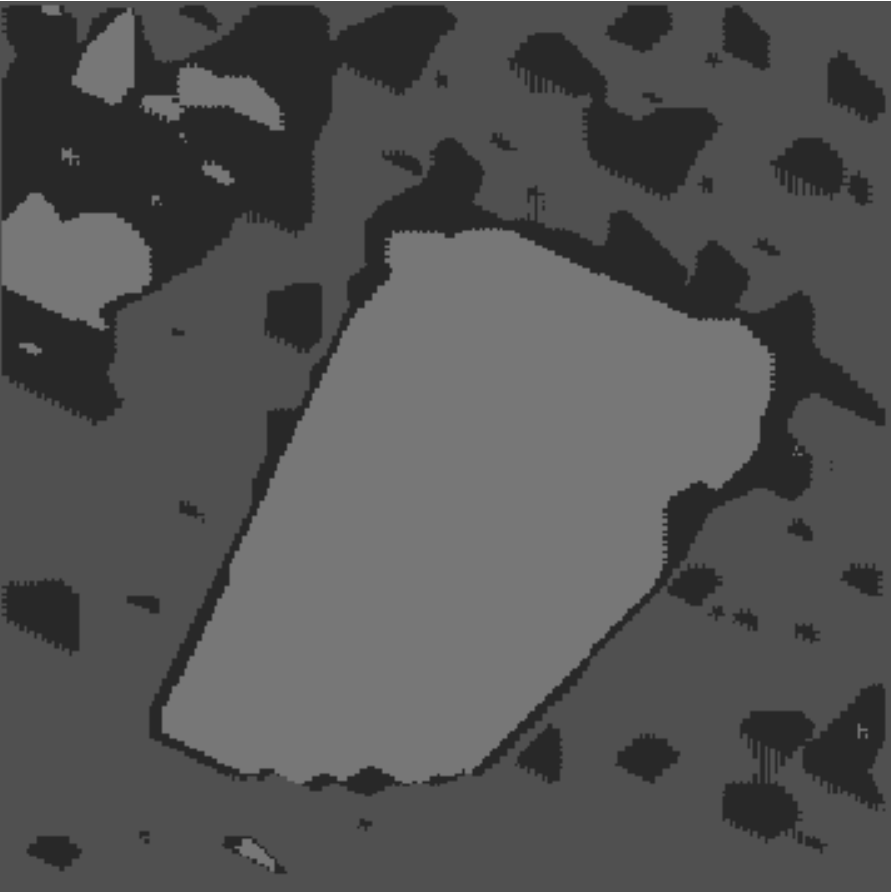}}
 \caption{Training by region of interest: real image}\label{fig:visual}
\end{sidewaysfigure}

Figure~\ref{fig:visual} compares the performance of the proposed stack filter with respect to two widely used SAR filters: Lee and Frost.
Figure~\ref{fig:realimage} presents the original data, and the regions of interest used for estimating the Boolean function; the data are from the ESAR sensor and show two distinct agricultural areas with possibly a raw of trees (the high return strip).
The data are from the HV polarization, L-band with about \unit[$1$]{m}$\times$\unit[$1.5$]{m} pixel size and of \unit[$3$]{m}$\times$\unit[$2.2$]{m} spatial resolution.
In this case, again, the mean on each region was used as the `ideal' image.
Figures~\ref{fig:filfrost}, \ref{fig:fillee}, \ref{fig:stackfilter1} and~\ref{fig:stackfilter95} present the result of applying the Frost, Lee and Stack filters (one and $22$ iterations; this last choice, again, obtained by visual inspection) to the original SAR data.
The right side of previous figures present the corresponding GMLC.
The stack filter produces better results than classical despeckling techniques.

Table~\ref{tab:confusion223} presents the main results from the confusion matrices of all the GMLC, including the results presented in~\cite{Buemi:Sibgrapi:2007} which used the classical stack filter estimation with whole images.
It shows the percentage of pixels that was labeled by the user as from region $R_i$ that was correctly classified as belonging to region $R_i$, for $1\leq i\leq 3$.
``None'' denotes the results on the original, unfiltered, data, ``Sample Stack $k$'' denotes our proposal of building stack filters with samples, applied $k$ times, ``Stack $k$'' the classical construction applied $k$ times, and ``Frost'' and ``Lee'' the classical speckle reduction filters.
It is clear the superior performance of stack filters (both classical and by training) over speckle filters, though the stack filter by training requires more than a single iteration to outperform the last ones.
Stack filters by training require about two orders of time less than classical stack filters to be built, and they produce comparable results.
Using regions of interest is, therefore, a competitive approach for the definition of this kind of filters.
\begin{table}[h!]%[hbt]
\caption{Statistics from the confusion matrices}\label{tab:confusion223}
    \centering
%\resizebox*{7cm}{4cm}{
    \begin{tabular}{ c rrr }\toprule
            Filter & $R_1$/$R_1$ & $R_2$/$R_2$ & $R_3$/$R_3$   \\ \midrule
                   None & $13.40$ & $48.16$ & $88.90$\\ \midrule
	            Sample Stack 1   & $9.38$ & $65.00$ & $93.19$\\
        	    Sample Stack 22  & $\textbf{63.52}$ &$\textbf{74.87}$ & $\textbf{96.5}$\\ \midrule
                    Stack 1  & $14.35$  & $64.65$ & $90.86$\\
                    Stack 40 & $62.81$  & $89.09$ & $94.11$\\
                    Stack 95 & $\textbf{63.01}$ & $\textbf{93.20}$ &$\textbf{94.04}$\\ \midrule
                    Frost &$16.55$ &$55.54$ &$90.17$\\
                    Lee   &$16.38$ &$52.72$ &$89.21$\\ \bottomrule
        \end{tabular}
%}
\end{table}

\subsection{Contrast preservation}

We use a second phantom of strips and points to assess contrast and edge preservation. 
It consists of  an image of size $256\times256$ pixels with two regions: one formed by strips of several widths on a background. 
The data in the former are distributed according to a $\mathcal G^0(\alpha_1,\gamma^*_{\alpha,1},1)$ law, and the latter obey a $\mathcal G^0(\alpha_2,\gamma^*_{\alpha_2,1},1)$ distribution, where $\gamma^*_{\alpha,n}$ is the scale, $\alpha \in \{-10,\dots,-1 \}$ is the roughness parameter and the number of looks is held constant in the noisiest case, namely $n =1$.

The factor used in this experiment was the contrast between light areas (strips and points) and background, measured as
\begin{equation*}
\frac{|\mu_1 - \mu_2|}{\sqrt{\sigma_1^2 + \sigma_2^2} },
\label{another_contrast_measure}
\end{equation*}
which considers the mean and the standard deviation of each region. 
These values, which depend on the three parameters of the $\mathcal{G}^0$ distribution can be computed using the expression of moments given in equation~\eqref{eq:GI0Moments}.
\citet{NascimentoCintraFreryIEEETGARS} present other contrast measures based on stochastic divergences between distributions.

\begin{sidewaysfigure}[htb]
  \begin{center}
\subfigure[Phantom and simulated with $\alpha_1 = -1$ and $\alpha_2 = -5$. \label{fig:phantom}]{\includegraphics[width=.30 \linewidth]{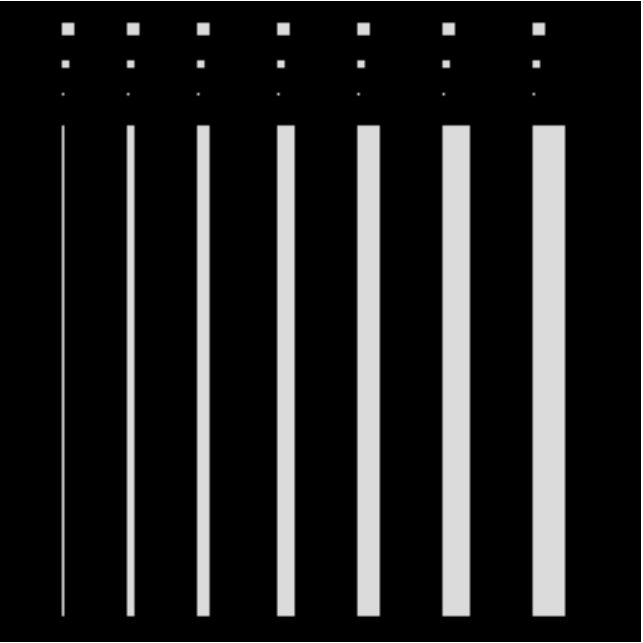}
\includegraphics[width=.30 \linewidth]{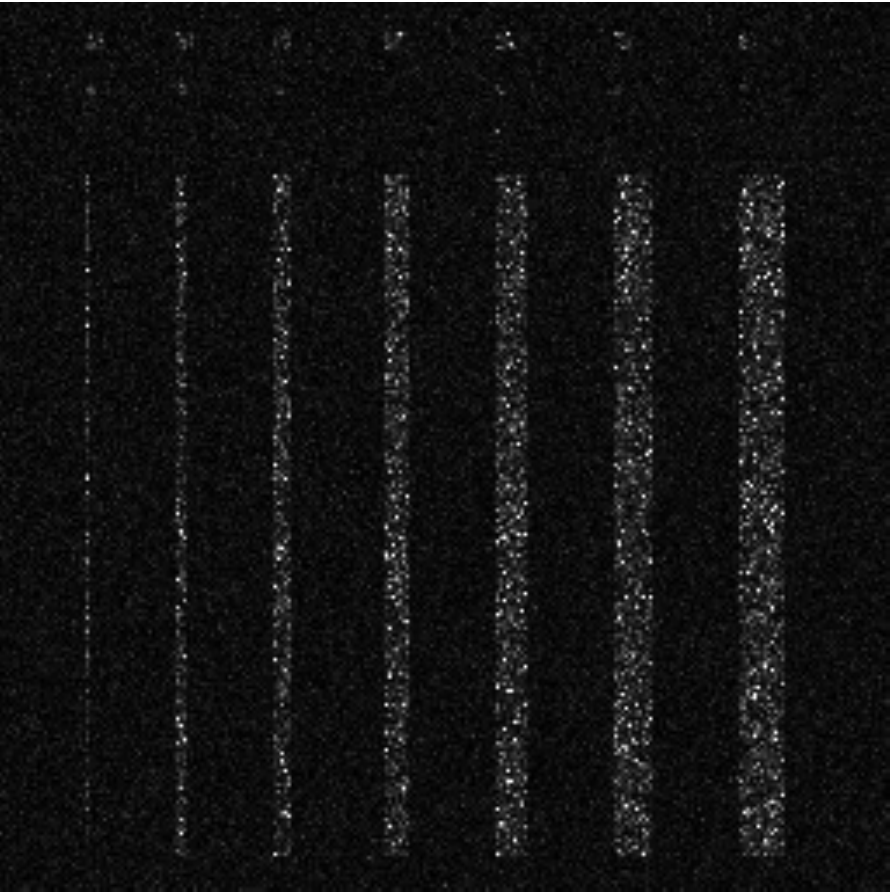}}
\\
\subfigure[Stack, Lee and Frost on data with $\alpha_1 = -1.$ and $\alpha_2 = -5. $1-5 \label{fig:stack-lee-1-5}]{ \includegraphics[width=.30 \linewidth]{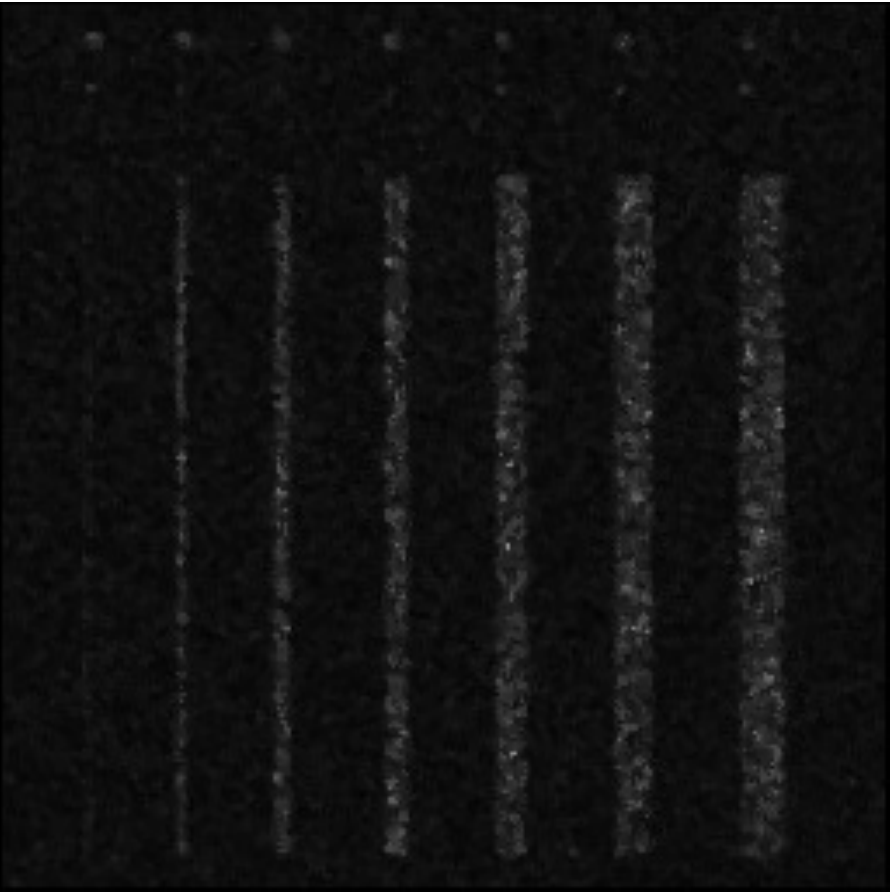}
\includegraphics[width=.30 \linewidth]{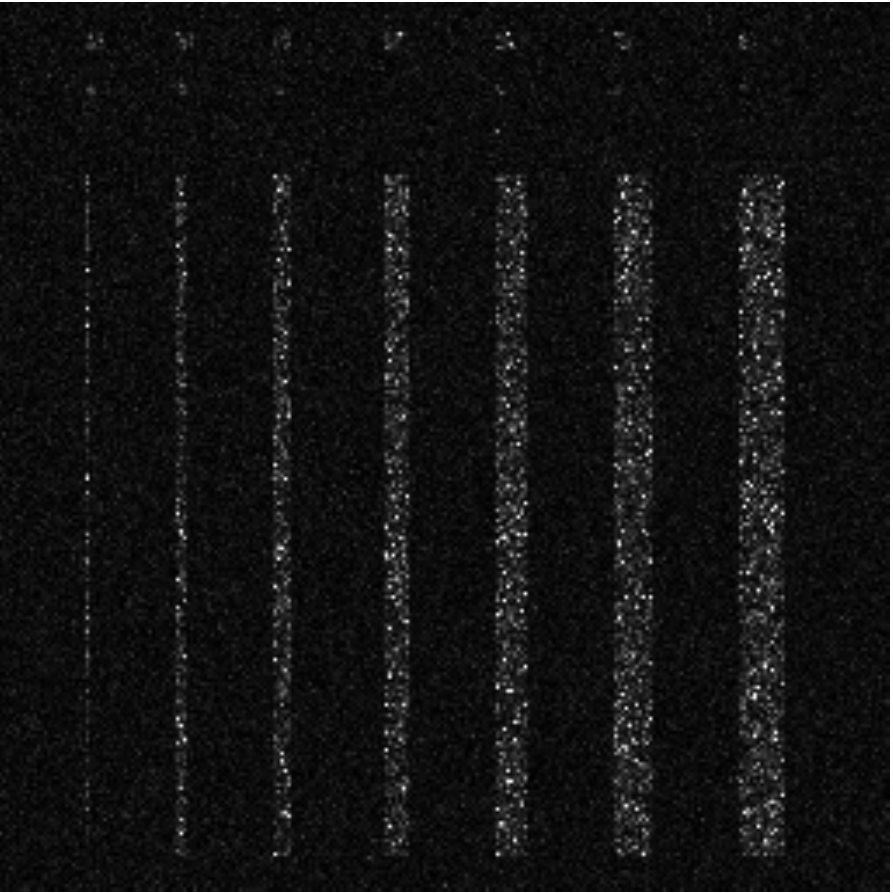} \includegraphics[width=.3 \linewidth]{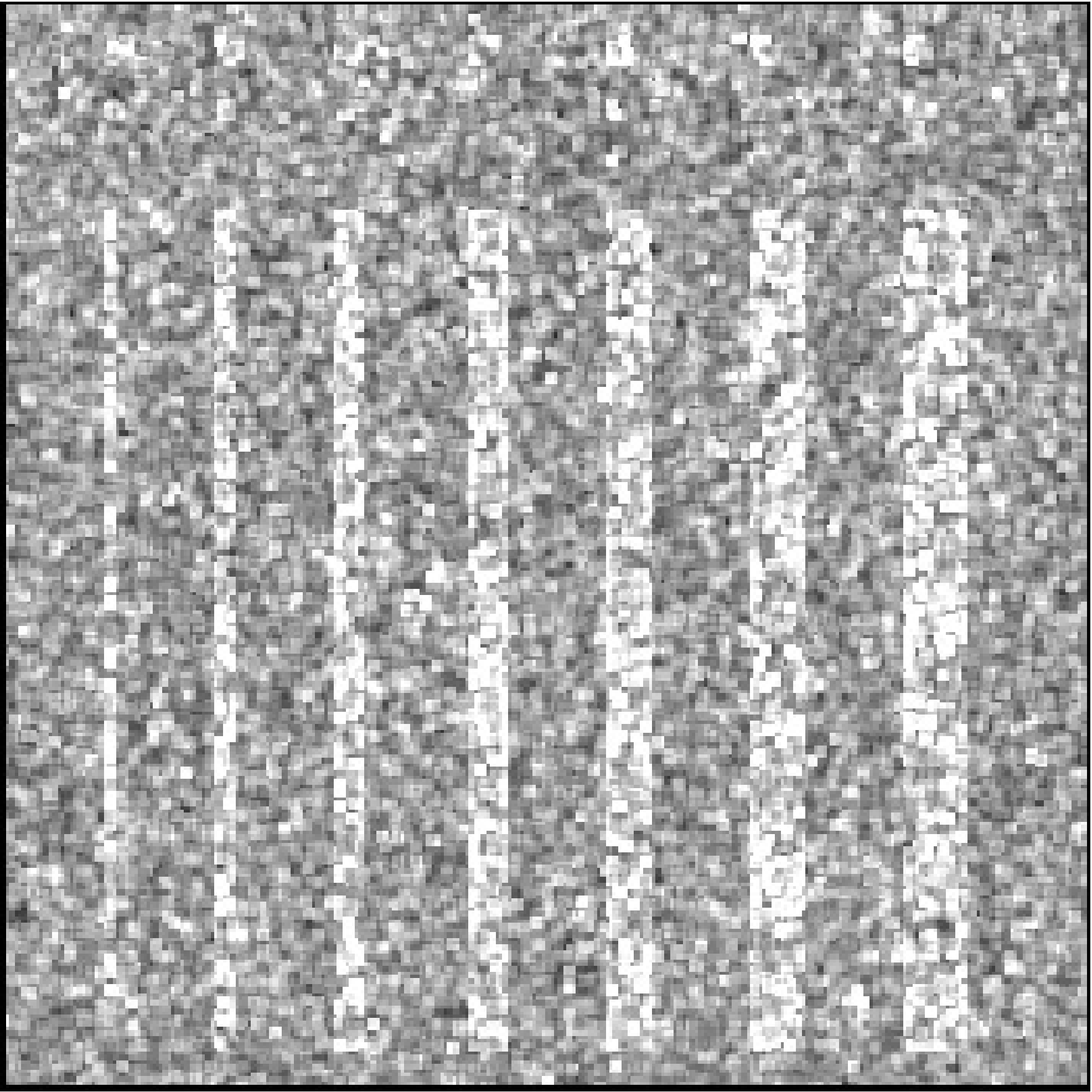}}
\\
%\subfigure[Stack, Lee and Frost on data with $\alpha_1 = -7.$ and $\alpha_2 = -3. $\label{fig:speckle-stack-lee-7-3}]{
%\includegraphics[width=.30 \linewidth]{speckle-7-3Eq.eps}
%\includegraphics[width=.30 \linewidth]{Lee-7-3Eq.eps}
%\includegraphics[width=.30 \linewidth]{frost73.eps}} 
 \end{center}
 \caption{Phantom, simulated data and Filtered with stack, Lee and Frost}
\label{fig:phantom_simulated}
\end{sidewaysfigure}

An ideal filter reduces the noise without affecting the contrast.
A blurry filter will smudge the edges, mixing classes and reducing the contrast.
Since the phantom presented in Figure~\ref{fig:phantom} consists of thin light strips and isolated points on a dark background, the strips and points will tend to disappear if the filter introduces blur.

Table~\ref{errorcontraste} presents the relative contrast error between the theoretical and observed values; the latter is the mean over one hundred replications.

%%%MEB Agregue comentada la columna Kuan, su imagen respectiva en folder <pdf>

\begin{table}[hbt]
\centering
\caption{Relative contrast error introduced by the filters}
\begin{tabular}{c l l l l} \toprule
$\alpha_1,\alpha_2$ & Contrast &Stack &Lee & Frost\\ \midrule% & Kuan\\ \midrule
$-1,-5$ & $0.0033$ & $0.0096$ & $0.0094$	& $0.0883$\\% & $0.1080$\\
$-2,-8$ & $0.0033$ & $0.0049$ & $0.0093$	& $0.2470$\\% & $0.2097$\\
$-6,-2$ & $0.0034$ & $0.0088$ & $0.0097$	& $0.2095$\\% & $0.1782$\\
%$-7,-3$ & 0.00337  &-0.1075074184&0.0095454545	&-3529.99\\
%$-9-3$ & 0.00337 &-0.1083976261	&0.0095864662	&-3234.3520178042\\
%$-10-2$& 0.00337 &-0.0161127596	&0.0061590909	&-1562.0819881306\\	
\bottomrule
\end{tabular}
\label{errorcontraste}
\end{table}

The Frost filter presents the best performance, but it also alters the contrast between regions as can be seen in Figure~\ref{fig:phantom_simulated}.
The stack filter outperforms by a small margin the Lee filter in two out of three cases, but both are much better than the Frost filter regarding contrast preservation.
Visual inspection confirms previous results: the stack filter is more effective at reducing speckle than the other two techniques, as can be seen in Figure~\ref{fig:phantom_simulated}.

\section{Conclusions} \label{sec:Conclusiones}
In this work, the effect of adaptive stack filtering on SAR images was assessed. Two viewpoints were considered: a classification performance viewpoint and a quality perception viewpoint. For the first approach, the Frost and Lee filters were compared with the iterated stack filter using a metric extracted from the confusion matrix. A real SAR image was used in this case.
For the second approach, a Monte Carlo experience was carried out in which 1-look synthetic SAR, i.e., the noisiest images, were generated. In this case, the Lee filter and a one pass stack filter were compared for various degrees of contrast. The $\beta$ and the $Q$ indexes were used as measures of perceptual quality.
The results of the $\beta$ index shows that the stack filter performs better in cases of high contrast.
The results of the $Q$ index show slightly better performance of the stack filter over the Lee filter. This quality assessment is not conclusive but indicates the potential of stack filters in SAR image processing for visual analysis.
The classification results and the quality perception results suggest that stack filters are promising tools in SAR image processing and analysis.

The results obtained using the contrast measure given by formula~\ref{another_contrast_measure} show a comparatively good performance of the stack Filter.
The system was developed in Matlab, and the code is available from the first author upon request.

\bibliography{bib-maelena}
\bibliographystyle{elsarticle-harv}
% 
% 
% \bibliographystyle{IEEEbib}
% \bibliography{bib-maelena}
\end{document}